\documentclass[]{fairmeta}

\usepackage{multirow}
\usepackage{graphicx}
\usepackage{array}
\usepackage{hyperref}
\usepackage{amsmath,amssymb,amsfonts}
\usepackage{cleveref}
\usepackage{subcaption}
\usepackage[numbers]{natbib}
\usepackage{booktabs}
\usepackage{enumitem}
\usepackage{fontawesome5}
\usepackage{algorithm}
\usepackage{algorithmic}

\newcolumntype{L}[1]{>{\raggedright\arraybackslash}m{#1}}
\newcolumntype{C}[1]{>{\centering\arraybackslash}m{#1}}

\newcommand{\commentout}[1]{}
\renewcommand{\paragraph}[1]{\noindent\textbf{#1}\hspace*{1em}}
\setlist[itemize]{leftmargin=15pt}

\title{OPD-V: Visual On-Policy Self-Distillation with Modality Balance}

\author[2,3]{Aniri}
\author[1,2,3]{Jinhe Bi}
\author[4]{Peng Liao}
\author[2]{Zengjie Jin}
\author[2,3]{Volker Tresp}
\author[1]{Fei Shen}
\author[2,3]{Yunpu Ma\,\faEnvelope}
\author[1]{Tat-Seng Chua}

\affiliation[1]{National University of Singapore}
\affiliation[2]{Ludwig Maximilian University of Munich}
\affiliation[3]{Munich Center for Machine Learning}
\affiliation[4]{Sun Yat-sen University}

\vspace{0.5em}
\contribution[]{\href{https://github.com/aniri15/OPD-V}{\faGithub~ GitHub} \quad
\href{https://huggingface.co/aaniri/OPD-V-Qwen3-VL-8B-Instruct}{\faCube~ Qwen3-VL-8B} \quad
\href{https://huggingface.co/aaniri/OPD-V-Qwen3-VL-4B-Instruct}{\faCube~ Qwen3-VL-4B} \quad
\href{https://huggingface.co/aaniri/OPD-V-Qwen3.5-9B}{\faCube~ Qwen3.5-9B} \quad
\href{https://huggingface.co/aaniri/OPD-V-Qwen3.5-4B}{\faCube~ Qwen3.5-4B}}

\abstract{
On-Policy Self-Distillation (OPSD) has become a standard post-training approach for improving visual reasoning in multimodal large language models (MLLMs). Existing methods draw privileged information from diverse input sources to guide self-distillation. Yet these designs overlook Modality Imbalance, a challenge inherent to MLLM reasoning. When textual information dominates generation, the model cannot fully integrate its multimodal input. Consequently, carefully designed privileged information remains underused, limiting the effectiveness of OPSD. To examine this limitation, we construct a Positive Teacher with the Zoom-In Image and a Negative Teacher with the Mask Image, which exhibit different degrees of Modality Imbalance. Changes in their reasoning correctness and token logits reveal that Modality Balance can itself serve as privileged information. Motivated by this finding, we introduce OPD-V, a visual OPSD paradigm that instantiates such information through the Positive Teacher and Negative Teacher. Positive Modality-Balance Logits Margins define a Modality-Balance Trust Region that selects the on-policy tokens used for self-distillation. Experiments across 6 benchmarks, 4 MLLM backbones, and 5 post-training methods show that OPD-V consistently improves reasoning performance while reducing training cost.
}
\date{\today}
\correspondence{\email{bijinhe@outlook.com}, \email{cognitive.yunpu@gmail.com}}

\begin{document}
\thispagestyle{firstheader}
\maketitle

\section{Introduction}

Multimodal large language models (MLLMs) support a broad range of visual understanding and reasoning tasks \cite{zhao2026nl2codestructuredsurveymultimodal,ma-etal-2026-self,zhang2023spot,huang2025loongsynthesizelongchainofthoughts}. On-Policy Self-Distillation (OPSD) has recently become a standard post-training approach for further improving their visual reasoning. OPSD samples trajectories from the current student and uses a copy of the same model to provide dense targets at student-visited prefixes \cite{agarwal2024policy,lu2025onpolicydistillation,zhao2026self}. This model copy receives privileged information available only during training, removing the need for a separately trained external teacher.

The quality of these targets depends on how the privileged information is constructed. A common strategy provides verified textual solutions or reference answers \cite{zhao2026self,yu2026dopddualonpolicydistillation,bi2026reflectrllearninggoldennegative}. Recent visual OPSD methods instead derive privileged information from additional or transformed visual inputs. Vision-OPD conditions the teacher on an evidence-centered crop, whereas Visual-OPSD uses privileged visual thoughts \cite{yuan2026visionopd,li2026visualopsd}. VA-OPD and VCSD create teacher contrasts by changing access to visual content \cite{liu2026visualadvantage,liang2026visualcontrastive}. Although these methods construct the privileged signal in different ways, they do not account for Modality Imbalance, a challenge inherent to MLLM reasoning.

MLLMs generate each response from visual inputs, textual instructions, and autoregressive context. This joint conditioning creates a persistent asymmetry because strong textual priors can dominate generation even when the task depends on the image \cite{bi-etal-2025-llava,park2025generalizing,zhang2025evaluating,zhang2026instruction}. We define this tendency as \emph{Modality Imbalance}: the model relies disproportionately on textual context instead of integrating the multimodal input. Carefully designed privileged information can enrich the teacher input without revealing how the visual and textual modalities contribute to each prediction. Consequently, the model can underuse this information, limiting the effectiveness of OPSD. 
This limitation raises the central question of our work: 

\textit{How can privileged information be designed to remain effective for MLLM OPSD under Modality Imbalance?}

\textbf{The Present Work:}
We address this question by treating \emph{Modality Balance itself} as privileged information for OPSD and introducing \textbf{OPD-V}. Unlike conventional privileged information like textual solutions or visual inputs, Modality Balance is not an explicit input that can be directly appended to the teacher context. It instead describes the model's internal allocation of visual and textual information during generation. The central challenge is therefore to express this internal condition through explicit teacher inputs and convert it into token-level supervision for OPSD.

\begin{figure*}
    \centering
    \makebox[\textwidth][c]{\includegraphics[width=1.08\textwidth]{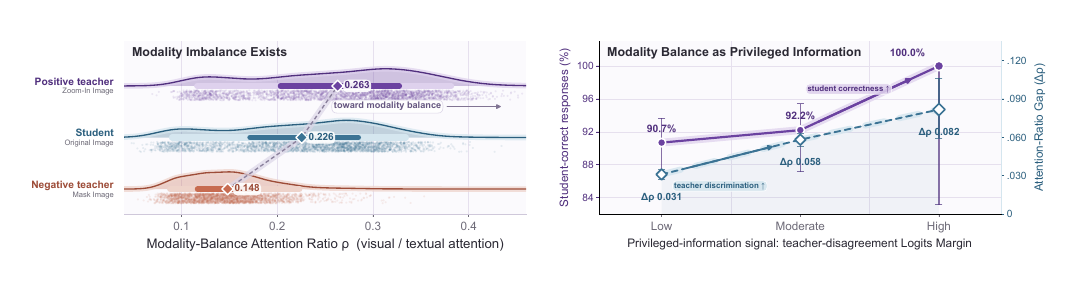}}
    \caption{Modality Balance as privileged information, evaluated on 5K samples drawn from multiple domains. Left: the Modality-Balance Attention Ratio is lowest for the Negative Teacher with the Mask Image, intermediate for the student with the Original Image, and highest for the Positive Teacher with the Zoom-In Image. These matched conditions expose distinct degrees of Modality Imbalance. Right: as the Modality-Balance Logits Margin increases, the gap between the Positive and Negative Teacher attention ratios widens, accompanied by higher student correctness. For each margin interval \(b\), the blue curve reports \(\Delta\rho_b=|b|^{-1}\sum_{i\in b}[\rho(c^+;y_i)-\rho(c^-;y_i)]\), where \(y_i\) is the response trajectory for sample \(i\), and \(c^+\) and \(c^-\) denote the Zoom-In Image and Mask Image conditions, respectively.}
    \label{fig:modality_balance_privilege}
\end{figure*}
Figure~\ref{fig:modality_balance_privilege} demonstrates this conversion through matched student and teacher image conditions. We construct a Positive Teacher with the Zoom-In Image and a Negative Teacher with the Mask Image, while the student receives the Original Image. The Modality-Balance Attention Ratio measures visual attention relative to textual attention under each condition. As shown in the left panel, the ratio is lowest for the Negative Teacher, intermediate for the student, and highest for the Positive Teacher. This ordering shows that the matched image conditions expose distinct degrees of Modality Imbalance through explicit differences in the model input.

The right panel further examines how these differences affect teacher scoring. The Modality-Balance Logits Margin measures how differently the Positive Teacher and Negative Teacher score the student response. From low to high margin intervals, the gap between their Modality-Balance Attention Ratios widens, while student correctness increases. Teachers with different levels of Modality Balance thus provide a directional scoring distinction that can serve as privileged information for OPSD.

\begin{figure*}
    \centering
    \includegraphics[width=1\linewidth]{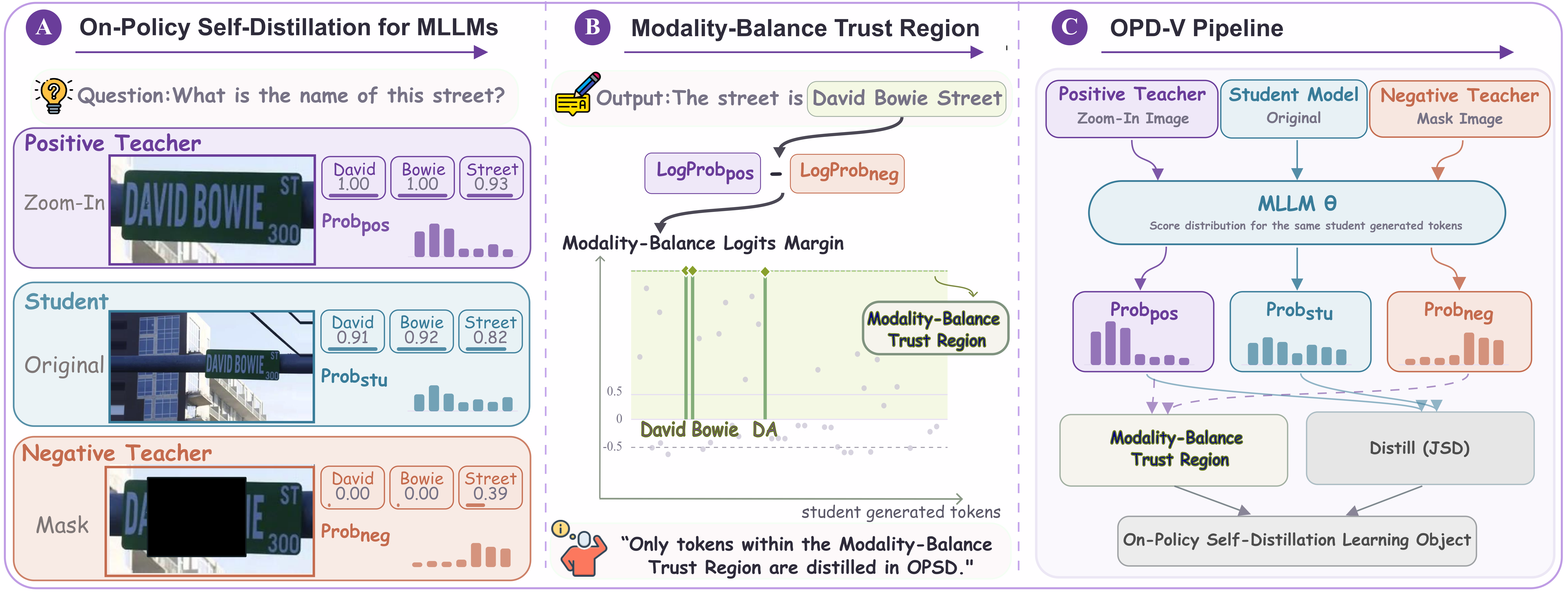}
    \caption{Overview of OPD-V. (A) For the same student-generated response, the Positive Teacher, student, and Negative Teacher score each on-policy token under the Zoom-In Image, Original Image, and Mask Image, respectively. (B) The tokenwise Modality-Balance Logits Margin compares the log probabilities assigned by the Positive Teacher and Negative Teacher; tokens with positive margins form the Modality-Balance Trust Region. (C) OPD-V applies Jensen--Shannon distillation from the Positive Teacher distribution to the student only within this region.}
    \label{fig:opdv_overview}
\end{figure*}

Building on this finding, we introduce \textbf{OPD-V}, a visual OPSD paradigm that uses Modality Balance as privileged information. Figure~\ref{fig:opdv_overview} summarizes how OPD-V implements this paradigm during training. The student receives the Original Image, the Positive Teacher receives the Zoom-In Image, and the Negative Teacher receives the Mask Image. All three distributions score the same student-generated tokens under matched textual context and on-policy prefixes. The Modality-Balance Logits Margin then compares the scores assigned by the Positive Teacher and Negative Teacher. Positive Modality-Balance Logits Margins define the Modality-Balance Trust Region used for self-distillation. Within this region, OPD-V applies Jensen--Shannon distillation from the Positive Teacher distribution to the student.

We evaluate OPD-V across 6 benchmarks, 4 MLLM backbones, and 5 post-training methods. These benchmarks span visually demanding settings and broader multimodal reasoning tasks across image scale, resolution, language, and real-world context. OPD-V consistently improves reasoning performance across the evaluated backbones and settings. Compared with standard OPSD, it also reduces step latency on both 4B and 9B backbones. These results show that OPD-V uses Modality Balance as privileged information to improve reasoning accuracy while reducing training cost.

Our contributions are summarized as follows:
\begin{itemize}
    \item We identify Modality Imbalance as a limitation that restricts the effectiveness of privileged information in MLLM OPSD.
    \item We show that Modality Balance can serve as privileged information and propose OPD-V, which instantiates it through the Positive Teacher, Negative Teacher, and Modality-Balance Trust Region.
    \item We demonstrate consistent improvements across representative OPSD methods while reducing training cost.
\end{itemize}

\section{Preliminaries}

\subsection{Task Formulation}\label{sec:task_formulation}

We consider supervised post-training for multimodal reasoning. Each example contains an Original Image \(I\), a textual query \(x\), and a verified target \(a^\star\) available only during training. The Original Image may include the target-region cue, and the corresponding visual focus instruction is included in the textual query. The trainable MLLM \(p_\theta\) receives only the Original Image and textual query, and samples the on-policy trajectory
\begin{equation}
    y=(y_1,\ldots,y_T)\sim p_\theta(\,\cdot\mid I,x).
    \label{eq:student_rollout}
\end{equation}
Let \(y_{<t}=(y_1,\ldots,y_{t-1})\) be the student-generated prefix before token \(y_t\), and let \(\mathcal{T}_y=\{1,\ldots,T\}\) be the corresponding token-index set. At position \(t\), the student distribution is
\begin{equation}
    p_{\theta,t}(\,\cdot\,)
    :=p_\theta(\,\cdot\mid I,x,y_{<t}).
    \label{eq:student_token_distribution}
\end{equation}
The verified target \(a^\star\) is privileged information used only to construct training-time teacher supervision.

\subsection{On-Policy Self-Distillation for MLLMs}

On-policy self-distillation (OPSD) samples \(y\) from the current student and evaluates each student-generated prefix with a detached copy of the same MLLM \cite{agarwal2024policy,lu2025onpolicydistillation,zhao2026self}. Let \(\bar{\theta}\) denote the detached teacher parameters. At token position \(t\), the teacher receives the verified target \(a^\star\) and scores the same student prefix:
\begin{equation}
    q_t^{\mathrm{priv}}(\,\cdot\,)
    :=
    p_{\bar{\theta}}(\,\cdot\mid I,x,a^\star,y_{<t}).
    \label{eq:opsd_teacher_distribution}
\end{equation}
Vanilla OPSD distills this privileged teacher distribution into the student along its on-policy trajectory:
\begin{equation}
    \begin{aligned}
    \mathcal{L}_{\mathrm{OPSD}}(\theta)
    &=
    \mathbb{E}_{y\sim p_\theta(\cdot\mid I,x)}
    \Bigg[
        \frac{1}{|\mathcal{T}_y|}
        \\
    &\qquad\cdot
        \sum_{t\in\mathcal{T}_y}
        D_{\mathrm{dist}}\!\left(
            q_t^{\mathrm{priv}},p_{\theta,t}
        \right)
    \Bigg].
    \end{aligned}
    \label{eq:opsd_objective}
\end{equation}
Here \(D_{\mathrm{dist}}\) denotes the token-level distillation divergence, instantiated as \(D_{\mathrm{JS}}\) in our implementation. Since each \(y_{<t}\) is generated by the current student, supervision remains on-policy and gradients propagate only through \(p_{\theta,t}\).

\subsection{Modality Imbalance in MLLMs}

During OPSD, each token prediction is conditioned on the visual input, textual query, and autoregressive prefix. Modality Imbalance denotes the tendency of an MLLM to depend disproportionately on textual context relative to visual input. Consequently, a prediction can be mainly supported by language priors even when the query requires visual information \cite{park2025generalizing,zhang2025evaluating,zhang2026instruction}. To measure this tendency over a complete response, let \(\alpha_t^V(c)\) and \(\alpha_t^T(c)\) denote the visual and textual attention masses at position \(t\) under model condition \(c\). We define the Modality-Balance Attention Ratio as
\begin{equation}
    \rho(c;y)
    =
    \frac{\sum_{t\in\mathcal{T}_y}\alpha_t^V(c)}
    {\sum_{t\in\mathcal{T}_y}\alpha_t^T(c)+\epsilon}.
    \label{eq:modality_balance_attention_ratio}
\end{equation}
where \(\epsilon\) is a small constant for numerical stability. A larger \(\rho(c;y)\) indicates stronger visual attention relative to textual attention over the on-policy trajectory. This ratio measures Modality Imbalance without requiring equal numerical contributions from the two modalities.

\section{OPD-V}

\subsection{Overview}

After measuring Modality Imbalance with the Modality-Balance Attention Ratio, OPD-V turns Modality Balance into OPSD supervision. For an on-policy trajectory \(y\sim p_\theta(\cdot\mid I,x)\), the student first generates tokens from the Original Image. The Positive Teacher with the Zoom-In Image and the Negative Teacher with the Mask Image then score the same tokens under the matched textual query and prefix. Comparing their normalized logits gives the Modality-Balance Logits Margin. When this margin is positive, the token receives stronger support after the task-relevant region is magnified than after selected visual regions are masked, and is therefore placed in the Modality-Balance Trust Region. The objective distills the Positive Teacher distribution only within this selected region.

\subsection{Positive Teacher with the Zoom-In Image and Negative Teacher with the Mask Image}

The two teacher views are constructed sequentially. \(I^{\mathrm{zoom}}\) is obtained by cropping and magnifying the task-relevant region. The Mask Image is then generated as \(I^{\mathrm{mask}}=\operatorname{Mask}(I^{\mathrm{zoom}})\) by replacing a random rectangular region in the Zoom-In Image with black pixels. Both inputs are passed to the detached teacher with the same textual query and student-generated prefix:
\begin{equation}
    \begin{aligned}
    q_t^+(\cdot)
    &=
    p_{\bar{\theta}}(\cdot\mid I^{\mathrm{zoom}},x,y_{<t}),
    \\
    q_t^-(\cdot)
    &=
    p_{\bar{\theta}}(\cdot\mid I^{\mathrm{mask}},x,y_{<t}).
    \end{aligned}
    \label{eq:opdv_teacher_distributions}
\end{equation}
The Positive Teacher supplies the distillation target, while the Negative Teacher supplies the paired comparison for the same student-generated token \(y_t\).

\subsection{Modality-Balance Trust Region}

Since the two teacher distributions are evaluated at the same prefix, token selection can be based on their normalized logits. Let \(\ell_t^+(v)=\log q_t^+(v)\) and \(\ell_t^-(v)=\log q_t^-(v)\). For each on-policy token \(y_t\), the Modality-Balance Logits Margin is
\begin{equation}
    \delta_t^{\mathrm{MB}}
    =
    \ell_t^+(y_t)
    -
    \ell_t^-(y_t).
    \label{eq:opdv_mb_margin}
\end{equation}
A positive Logits Margin indicates that the same token receives more support from the Positive Teacher than from the Negative Teacher. These positions form the Modality-Balance Trust Region:
\begin{equation}
    \mathcal{R}_{\mathrm{MB}}(y)
    =
    \left\{
        t\in\mathcal{T}_y
        \;\middle|\;
        \delta_t^{\mathrm{MB}}>0
    \right\}.
    \label{eq:opdv_trust_region}
\end{equation}
Thus, \(\mathcal{R}_{\mathrm{MB}}(y)\) keeps the on-policy tokens whose support increases under the Zoom-In Image relative to the Mask Image.

\begin{figure*}[t!]
\centering
\includegraphics[width=\textwidth]{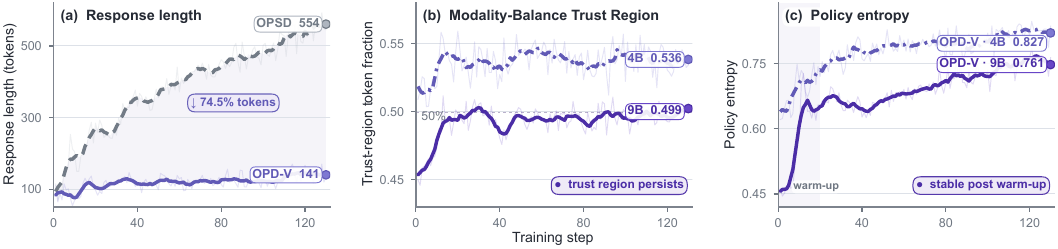}
\caption{Training dynamics on Qwen3.5 backbones. Thin traces show logged values, and bold curves show seven-step rolling means. (a) Response length on the 4B backbone, where OPD-V maintains concise responses while OPSD exhibits late-stage length growth. (b) Fraction of on-policy tokens within the Modality-Balance Trust Region for OPD-V at 4B and 9B scales. (c) Policy entropy for OPD-V at 4B and 9B scales, both of which remain stable after warm-up.}
\label{fig:training_dynamics}
\end{figure*}

\subsection{OPD-V Objective}

Within the Modality-Balance Trust Region, the Positive Teacher distribution serves as the Jensen--Shannon distillation target for the student distribution in Eq.~\eqref{eq:student_token_distribution}. Let \(r_t\in\{0,1\}\) be the valid-token indicator, where \(r_t=1\) means that position \(t\) is a generated response token included in distillation. This gives the OPD-V objective:
\begin{equation}
    \begin{aligned}
    \mathcal{L}_{\mathrm{OPD\text{-}V}}(\theta)
    &=
    \mathbb{E}\!\Bigg[
        \frac{1}{\sum_{t\in\mathcal{T}_y}r_t}
        \\
    &\qquad\cdot
        \sum_{t\in\mathcal{R}_{\mathrm{MB}}(y)}
        r_t
        \delta_t^{\mathrm{MB}}
        D_{\mathrm{JS}}\!\left(q_t^+,p_{\theta,t}\right)
    \Bigg].
    \end{aligned}
    \label{eq:opdv_loss}
\end{equation}
where \(D_{\mathrm{JS}}\) denotes Jensen--Shannon divergence. Because every \(t\in\mathcal{R}_{\mathrm{MB}}(y)\) has \(\delta_t^{\mathrm{MB}}>0\), the Modality-Balance Logits Margin scales the selected on-policy tokens during self-distillation.

\subsection{Computational Efficiency}

Compared with standard OPSD, OPD-V adds one Negative Teacher forward pass to score the same on-policy tokens under the Mask Image. The Modality-Balance Logits Margin and Modality-Balance Trust Region are computed tokenwise, while both teachers share the same EMA parameters; OPD-V therefore introduces no additional model parameters or student updates. In practice, shorter responses and more efficient teacher inputs offset this additional pass, as measured in Section~\ref{sec:exp_efficiency}.

\begin{table*}[t]
\centering
\footnotesize
\renewcommand{\arraystretch}{1.08}
\setlength{\tabcolsep}{0.8pt}
\begin{NiceTabular}{@{}L{0.205\textwidth}C{0.055\textwidth}C{0.095\textwidth}C{0.105\textwidth}C{0.09\textwidth}C{0.09\textwidth}C{0.095\textwidth}C{0.095\textwidth}C{0.115\textwidth}@{}}
\CodeBefore
    \rowcolor{metabg!10}{27}
    \rowcolor{metabg!10}{28}
\Body
\toprule
\multicolumn{1}{c}{\textbf{Model}} & \multicolumn{1}{c}{\textbf{Params}} & \multicolumn{1}{c}{\textbf{V* Bench}} & \multicolumn{1}{c}{\textbf{ZoomBench}} & \multicolumn{2}{c}{\textbf{HR-Bench}} & \multicolumn{2}{c}{\textbf{MME-RealWorld}} & \multicolumn{1}{c}{\textbf{Average}} \\
\cmidrule(lr){5-6} \cmidrule(lr){7-8}
\multicolumn{1}{c}{} & \multicolumn{1}{c}{} & \multicolumn{1}{c}{} & \multicolumn{1}{c}{} & \multicolumn{1}{c}{\textbf{4K}} & \multicolumn{1}{c}{\textbf{8K}} & \multicolumn{1}{c}{\textbf{EN}} & \multicolumn{1}{c}{\textbf{CN}} & \multicolumn{1}{c}{} \\
\midrule
\multicolumn{9}{c}{\textbf{Closed-Source Models (Single Forward Pass)}} \\
\midrule
GPT-5.2 & - & 79.06 & 50.89 & 81.12 & 78.38 & 72.60 & 68.80 & 71.81 \\
GPT-5.4 & - & 76.96 & 52.66 & 84.00 & 77.88 & 74.20 & 70.93 & 72.77 \\
Gemini-3.1-Pro & - & 87.96 & 61.18 & 89.63 & 86.88 & 76.53 & 73.31 & 79.25 \\
Gemini-3.5-Flash & - & 89.01 & 61.42 & 89.12 & 86.62 & 75.31 & 73.97 & 79.24 \\
\midrule
\multicolumn{9}{c}{\textbf{Open-Source Baselines (Single Forward Pass)}} \\
\midrule
DeepEyes & 7B & 85.86 & 46.51 & 75.13 & 72.63 & 64.10 & 64.09 & 68.05 \\
Thyme & 7B & 82.20 & 45.09 & 77.00 & 72.00 & 64.80 & 64.59 & 67.61 \\
DeepEyesV2 & 7B & 81.68 & 44.97 & 77.88 & 73.75 & 64.90 & 65.07 & 68.04 \\
SenseNova-MARS & 8B & 92.15 & 47.81 & 83.13 & 78.38 & 67.90 & 68.90 & 73.05 \\
MiMo-VL-RL & 7B & 83.25 & 45.68 & 73.50 & 69.38 & 62.73 & 55.89 & 65.07 \\
ZwZ & 8B & 87.96 & 56.69 & 83.63 & 81.75 & 66.57 & 68.09 & 74.12 \\
MiniCPM-V-4.5 & 9B & 70.68 & 42.60 & 69.63 & 61.50 & 62.65 & 61.64 & 61.45 \\
GLM-4.6V & 106B & 86.91 & 50.06 & 82.13 & 78.88 & 65.57 & 65.62 & 71.53 \\
Qwen3-VL-Instruct & 235B & 91.10 & 56.09 & 86.13 & 80.38 & 71.74 & 69.04 & 75.75 \\
Qwen3.5 & 397B & 87.96 & 57.16 & 89.38 & 85.50 & 74.82 & 69.82 & 77.44 \\
Kimi-K2.6 & 1T & 88.48 & 53.14 & 81.88 & 78.00 & 69.22 & 66.13 & 72.81 \\
\midrule
\multicolumn{9}{c}{\textbf{Qwen3.5 (4B) Distillation Series \& Ours}} \\
\midrule
Qwen3.5 & 4B & 80.63 & 49.82 & 73.50 & 67.50 & 55.78 & 58.56 & 64.30 \\
\quad \textit{+ SFT} & 4B & 83.25 & 42.60 & 76.50 & 72.12 & 64.10 & 62.50 & 66.85 \\
\quad \textit{+ GRPO} & 4B & 85.34 & 48.20 & 78.25 & 74.50 & 66.50 & 65.10 & 69.65 \\
\quad \textit{+ OPSD} & 4B & 84.29 & 57.51 & 77.25 & 72.88 & 61.20 & 62.40 & 69.26 \\
\quad \textit{+ Vision-OPD} & 4B & 89.01 & 65.33 & 82.25 & 81.25 & 74.50 & 70.26 & 77.10 \\
\quad \textit{+ VA-OPD} & 4B & 86.91 & 51.20 & 80.12 & 77.50 & 68.50 & 67.10 & 71.88 \\
\multirow{2}{*}{\quad \textit{+ \textbf{OPD-V (Ours)}}} & \multirow{2}{*}{4B} & \textbf{94.76} & \textbf{65.92} & \textbf{85.75} & \textbf{83.25} & \textbf{76.31} & \textbf{74.06} & \textbf{80.01} \\[-0.55ex]
& & \tiny(+14.1\%) & \tiny(+16.1\%) & \tiny(+12.3\%) & \tiny(+15.8\%) & \tiny(+20.5\%) & \tiny(+15.5\%) & \tiny(+15.7\%) \\
\bottomrule
\end{NiceTabular}
\caption{Comparison with state-of-the-art MLLMs and matched Qwen3.5-4B post-training baselines across six multimodal benchmarks. All entries report accuracy (\%), and Average is computed over V* Bench, ZoomBench, HR-Bench at 4K and 8K resolutions, and the English and Chinese subsets of MME-RealWorld. Results for a subset of the baseline models are taken from Vision-OPD \cite{yuan2026visionopd}. The final group compares SFT, GRPO, OPSD, Vision-OPD, VA-OPD, and OPD-V using the same backbone and training data. Parenthetical values in the OPD-V row report absolute percentage-point gains over the Qwen3.5-4B base model; bold marks the best result within this matched post-training group.}
\label{tab:main_results}
\end{table*}

\begin{figure*}[t]
\centering
\includegraphics[width=\textwidth]{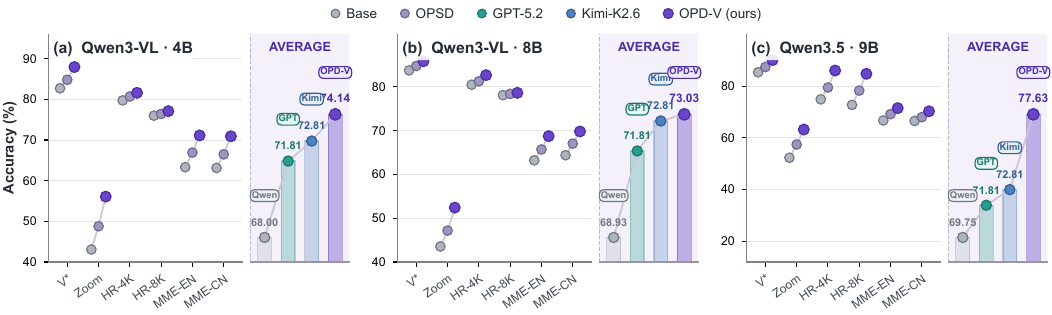}
\caption{Absolute accuracy of OPD-V across diverse model architectures (Qwen3-VL-Instruct 4B/8B and Qwen3.5 9B). Connected points compare the Base model, OPSD, and OPD-V on each benchmark. Model panels use independently zoomed y-axes to make within-backbone differences visible without altering the reported scores. The Average cards are independently scaled and annotate exact values for the Base model, GPT-5.2, Kimi-K2.6, and OPD-V.}
\label{fig:scalability}
\end{figure*}

\section{Experiments}\label{sec:exp}

In this section, we conduct extensive experiments to address the following research questions: (\textbf{RQ1: Overall Effectiveness}) How does OPD-V perform when compared to existing OPSD methods? (\textbf{RQ2: Computational Efficiency}) Does OPD-V introduce significant computational overhead? (\textbf{RQ3: Mechanistic Validity}) How sensitive is OPD-V to the proposed Dual Teacher Guidance Dynamics and the choice of Visual Transformation Strategy Variants?

\subsection{Experimental Settings}

\subsubsection{Model Training and Implementation Details}
We instantiate our proposed \textbf{OPD-V} framework across multiple vision-language backbones, including Qwen3.5-4B, Qwen3.5-9B, Qwen3-VL-4B-Instruct, and Qwen3-VL-8B-Instruct \cite{qwen3.5,bai2025qwen3vl}, utilizing a curated synthetic dataset of 6.2K visual reasoning samples from Vision-OPD \cite{yuan2026visionopd}. For policy alignment, we employ Jensen-Shannon Divergence (JSD) with $\beta = 0.5$ as the divergence objective. To overcome the high memory overhead inherent to full-vocabulary logit distillation \cite{zhao2026self}, we adopt a top-$K$ distillation strategy that evaluates only the top $K=100$ logits from the student model alongside their corresponding teacher probability distributions, supplemented by a tail-probability adjustment. In our experimental setup, tokens beyond the top 100 account for less than $1 \times 10^{-13}$ of the cumulative probability mass, making this top-$K$ approximation virtually lossless while significantly reducing memory requirements. The teacher parameters are maintained using an Exponential Moving Average (EMA) of the student weights. Throughout training, the maximum on-policy generation length is set to 1024 tokens, and models are trained for 1 epoch.

\subsubsection{Evaluation Benchmarks}
We evaluate OPD-V on six established benchmarks that connect visually demanding inputs to broader multimodal problem solving. Each requires models to extract task-relevant visual evidence and use it for question answering, contextual understanding, or reasoning. V* Bench \cite{wu2024v} tests the localization and recognition of small targets in high-resolution, visually crowded scenes, whereas ZoomBench \cite{wei2026zooming} examines whether models can recover subtle regional evidence from a full image across multiple fine-grained VQA dimensions. HR-Bench \cite{wang2025divide} extends image understanding to native 4K and 8K resolutions, where relevant information can be lost through conventional downsampling. MME-RealWorld \cite{zhang2024mme} provides the broadest setting, spanning diverse high-resolution tasks and real-world scenarios that require both detailed observation and contextual reasoning. Together, the suite provides a general multimodal evaluation across complementary challenges in evidence localization, scale, resolution, language, and real-world context.

\subsubsection{Baselines}
Following Vision-OPD \cite{yuan2026visionopd}, we compare \textbf{OPD-V} with two groups of baselines:
\begin{enumerate}
    \item \textbf{State-of-the-Art MLLMs:} In accordance with Table~\ref{tab:main_results}, we organize these baselines by model accessibility.
    (a) The closed-source group comprises the Google Gemini family, represented by Gemini-3.1-Pro \cite{google2026gemini3} and Gemini-3.5-Flash \cite{google2026gemini35}, and the OpenAI GPT family, represented by GPT-5.4 \cite{openai2026gpt54} and GPT-5.2 \cite{openai2025gpt52}.
    (b) The open-source group covers a broad range of model scales. Its compact 7B--9B systems include Thyme \cite{zhang2025thyme}, DeepEyes \cite{zheng2025deepeyes}, DeepEyesV2 \cite{hong2025deepeyesv2}, MiMo-VL-RL \cite{coreteam2025mimovltechnicalreport}, ZwZ \cite{wei2026zooming}, SenseNova-MARS \cite{chng2025sensenova}, and MiniCPM-V-4.5 \cite{yu2025minicpm}; the larger-scale systems comprise GLM-4.6V \cite{hong2025glm}, Kimi-K2.6 \cite{team2026kimi26}, Qwen3-VL-Instruct \cite{bai2025qwen3vl}, and Qwen3.5 \cite{qwen3.5}.

    \item \textbf{Alternative Optimization Strategies:} Using the same Qwen3.5-4B backbone and training data, we evaluate:
    (a) \emph{SFT}, supervised fine-tuning on the reference targets;
    (b) \emph{GRPO} \cite{shao2024deepseekmath}, reinforcement learning with binary ground-truth outcome verification;
    (c) \emph{OPSD} \cite{zhao2026self}, on-policy self-distillation conditioned on the reference targets;
    (d) \emph{Vision-OPD} \cite{yuan2026visionopd}, which constructs privileged information using evidence-centered crops; and
    (e) \emph{VA-OPD} \cite{liu2026visualadvantage}, which contrasts teacher predictions under different access to visual content.
\end{enumerate}
For this matched comparison, all post-training methods use the same 6.2K training set and the ground-truth target protocol from Vision-OPD \cite{yuan2026visionopd}. All models are trained and evaluated in non-thinking mode.

\begin{figure}[t]
\centering
\includegraphics[width=0.6\columnwidth]{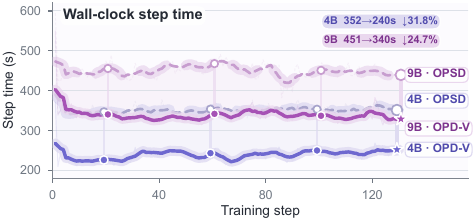}
\caption{Wall-clock step time for Qwen3.5 4B and 9B backbones. Pale traces show raw measurements and haloed curves show seven-step rolling means; dashed lines denote OPSD and solid lines denote OPD-V. Paired checkpoint connectors emphasize the time gap. Mean step latency drops from 352s to 240s (31.8\%) for 4B and from 451s to 340s (24.7\%) for 9B.}
\label{fig:training_time}
\end{figure}

\subsection{RQ1: Overall Effectiveness}
This analysis compares predictive performance and training dynamics across post-training strategies, model families, and parameter scales.

\paragraph{Takeaway 1: OPD-V improves performance across matched baselines, architectures, and scales.}
Table~\ref{tab:main_results} shows that OPD-V raises the Qwen3.5-4B average from 64.30\% to 80.01\%, an absolute gain of 15.71 percentage points. It also exceeds every matched post-training baseline: the strongest alternative, Vision-OPD, reaches 77.10\%, leaving a 2.91-point gap. OPD-V improves upon Vision-OPD on all six reported evaluations, so the aggregate gain spans the benchmark suite. The same pattern extends beyond Qwen3.5-4B. As shown in Figure~\ref{fig:scalability}, OPD-V improves Qwen3-VL-4B from 68.00\% to 74.14\%, Qwen3-VL-8B from 68.93\% to 73.03\%, and Qwen3.5-9B from 69.75\% to 77.63\%. These two model families differ in how multimodal information is represented and propagated. Qwen3-VL uses Interleaved-MRoPE and DeepStack to inject multi-level ViT features into the language backbone, whereas Qwen3.5 is pretrained with early vision--text fusion and adopts a 3:1 hybrid language backbone that interleaves Gated DeltaNet linear-attention layers with full-attention layers \cite{bai2025qwen3vl,qwen3.5}. The consistent improvements therefore span different mechanisms for visual feature integration, positional modeling, and token interaction, rather than only parameter scale.

Despite having only 4B parameters, the resulting Qwen3.5-4B model reaches 80.01\% average accuracy, outperforming every closed-source model listed in Table~\ref{tab:main_results}. It also surpasses substantially larger open-source systems, including the 397B Qwen3.5 model at 77.44\% and the 1T-parameter Kimi-K2.6 model at 72.81\%. These comparisons show that using Modality Balance as privileged information enables a compact model to achieve the strongest average performance among the reported single-forward-pass models across a general suite of multimodal tasks.

\paragraph{Takeaway 2: OPD-V maintains concise responses throughout training.}
Response length provides a behavioral view of reasoning efficiency, because shorter successful trajectories can reflect more direct problem solving \cite{Wu2025WhenMI,hassid2026dont}. Figure~\ref{fig:training_dynamics}(a) separates OPD-V from standard OPSD after the early training stage. The OPSD trajectory grows progressively longer, whereas OPD-V remains comparatively short and stable. Over the final ten recorded steps, the mean response length is 140.9 tokens for OPD-V and 553.8 tokens for OPSD, a 74.5\% reduction. OPD-V therefore pairs the performance improvement in Table~\ref{tab:main_results} with a compact response regime throughout training. The shorter responses also reduce token-level computation, which contributes directly to the efficiency result in RQ2.

\paragraph{Takeaway 3: OPD-V preserves stable policy entropy after warm-up.}
Figure~\ref{fig:training_dynamics}(c) shows closely bounded entropy trajectories for both Qwen3.5 scales after warm-up. Their final rolling means remain at 0.827 for 4B and 0.761 for 9B. The shared pattern indicates that OPD-V preserves a non-degenerate policy distribution across model sizes. Together with the sustained accuracy gains, this stability shows that OPD-V maintains an active learning signal throughout training.

\paragraph{Takeaway 4: The Modality-Balance Trust Region persists across training and model scales.}
Figure~\ref{fig:training_dynamics}(b) tracks the fraction of on-policy tokens whose Modality-Balance Logits Margin is positive. After warm-up, the trust-region fraction averages 53.8\% for Qwen3.5-4B and 49.6\% for Qwen3.5-9B, remaining near one half of each rollout. The region thus stays selective throughout optimization: the Positive Teacher and Negative Teacher consistently distinguish a subset of tokens for self-distillation. Similar fractions at 4B and 9B further show that this selection behavior is retained across scales.

\begin{figure}[t]
\centering
\includegraphics[width=0.6\columnwidth]{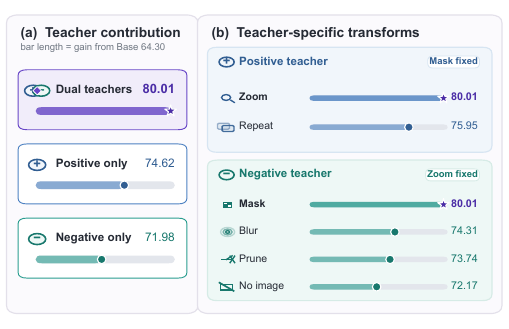}
\caption{Ablation studies on OPD-V components. Bar lengths encode gains from the 64.30\% base model, and labels report absolute accuracy. (a) Combining the Positive Teacher and Negative Teacher yields the highest accuracy (80.01\%). (b) Teacher-specific image-operation ablations: the Positive Teacher comparison holds Mask Image fixed and evaluates Repeat Image against Zoom-In Image, whereas the Negative Teacher comparison holds Zoom-In Image fixed and evaluates No Image, Prune, Blur, and Mask Image. Zoom-In Image and Mask Image form the strongest tested pair.}
\label{fig:ablation}
\end{figure}

\subsection{RQ2: Computational Efficiency}\label{sec:exp_efficiency}

\paragraph{Takeaway: OPD-V reduces step time through shorter rollouts and more efficient privileged-input construction.}
Figure~\ref{fig:training_time} shows that mean step latency decreases from 352\,s to 240\,s on Qwen3.5-4B and from 451\,s to 340\,s on Qwen3.5-9B, corresponding to reductions of 31.8\% and 24.7\%, respectively. Two stages account for this difference. First, the shorter OPD-V responses in Figure~\ref{fig:training_dynamics}(a) reduce the tokens processed during generation, student and teacher forward passes, and backpropagation. This effect is most visible on the 4B backbone, where the final-ten-step response length is 74.5\% lower than that of OPSD.
Second, the form of privileged information changes the cost of constructing and evaluating teacher inputs. Standard OPSD inserts the reference target into an answer-hint prompt and reconstructs the multimodal teacher sequence. OPD-V instead builds the Positive Teacher and Negative Teacher inputs by applying the Zoom-In Image and Mask Image operations while retaining the matched textual context. In the 9B timing logs, teacher-batch construction and preprocessing decrease from 79.2\,s for OPSD to 15.3\,s for OPD-V. Although OPD-V evaluates two teachers, their combined forward time is 27.1\,s, below the 40.4\,s required by the single OPSD teacher. Thus, the number of teacher passes alone does not determine training cost; response length and the construction of privileged inputs jointly explain why OPD-V trains faster.

\subsection{RQ3: Mechanistic Validity}\label{sec:ablation}

\paragraph{Takeaway 1: The Positive Teacher and Negative Teacher provide complementary supervision.}
Figure~\ref{fig:ablation}(a) shows that each teacher is effective in isolation. The Negative Teacher alone improves the base model from 64.30\% to 71.98\%, while the Positive Teacher alone reaches 74.62\%. Combining them yields 80.01\%, exceeding the stronger single-teacher variant by 5.39 percentage points and the Negative-Teacher-only variant by 8.03 points. This additional gain establishes the complementarity of the two teachers. The Positive Teacher supplies the target distribution, whereas the Negative Teacher determines which token positions enter the Modality-Balance Trust Region; the combined result aligns with these distinct roles in the OPD-V objective.

\paragraph{Takeaway 2: Zoom-In Image and Mask Image form the strongest tested operation pair.}
Figure~\ref{fig:ablation}(b) varies one teacher operation at a time. With Mask Image fixed for the Negative Teacher, replacing Zoom-In Image with Repeat Image lowers accuracy from 80.01\% to 75.95\%. With Zoom-In Image fixed for the Positive Teacher, Mask Image reaches 80.01\%, compared with 74.31\% for Blur, 73.74\% for Prune, and 72.17\% for No Image. These comparisons reveal two properties of the selected pair. Zoom-In Image changes the prominence of the task-relevant region, whereas repeating the Original Image leaves its visual composition unchanged. Mask Image modifies selected regions while preserving the remaining image context, whereas No Image removes that context entirely. Among the tested operations, the resulting Zoom-In--Mask produces the most effective Modality-Balance Trust Region and the highest final accuracy.

\section{Conclusion}
This work identifies Modality Balance as privileged information for OPSD and introduces OPD-V, which uses the Positive Teacher and Negative Teacher to define a Modality-Balance Trust Region. Across six benchmarks and four MLLM backbones, OPD-V consistently improves reasoning performance while reducing training cost.

\bibliographystyle{unsrtnat}
\bibliography{aaai2027}

@article{lu2025onpolicydistillation,
  author = {Kevin Lu and Thinking Machines Lab},
  title = {On-Policy Distillation},
  journal = {Thinking Machines Lab: Connectionism},
  year = {2025},
  note = {https://thinkingmachines.ai/blog/on-policy-distillation},
  doi = {10.64434/tml.20251026},
}

@article{shao2024deepseekmath,
  title={Deepseekmath: Pushing the limits of mathematical reasoning in open language models},
  author={Shao, Zhihong and Wang, Peiyi and Zhu, Qihao and Xu, Runxin and Song, Junxiao and Bi, Xiao and Zhang, Haowei and Zhang, Mingchuan and Li, YK and Wu, Yang and others},
  journal={arXiv preprint arXiv:2402.03300},
  year={2024}
}

@inproceedings{agarwal2024policy,
  title={On-policy distillation of language models: Learning from self-generated mistakes},
  author={Agarwal, Rishabh and Vieillard, Nino and Zhou, Yongchao and Stanczyk, Piotr and Ramos Garea, Sabela and Geist, Matthieu and Bachem, Olivier},
  booktitle={International Conference on Learning Representations},
  volume={2024},
  pages={21246--21263},
  year={2024}
}

@article{zhao2026self,
  title={Self-Distilled Reasoner: On-Policy Self-Distillation for Large Language Models},
  author={Zhao, Siyan and Xie, Zhihui and Liu, Mengchen and Huang, Jing and Pang, Guan and Chen, Feiyu and Grover, Aditya},
  journal={arXiv preprint arXiv:2601.18734},
  year={2026}
}

@article{hinton2015distilling,
  title={Distilling the knowledge in a neural network},
  author={Hinton, Geoffrey and Vinyals, Oriol and Dean, Jeff},
  journal={arXiv preprint arXiv:1503.02531},
  year={2015}
}

@inproceedings{kim2016sequence,
  title={Sequence-level knowledge distillation},
  author={Kim, Yoon and Rush, Alexander M},
  booktitle={Proceedings of the 2016 conference on empirical methods in natural language processing},
  pages={1317--1327},
  year={2016}
}

@article{wei2026zooming,
  title={Zooming without Zooming: Region-to-Image Distillation for Fine-Grained Multimodal Perception},
  author={Wei, Lai and He, Liangbo and Lan, Jun and Dong, Lingzhong and Cai, Yutong and Li, Siyuan and Zhu, Huijia and Wang, Weiqiang and Kong, Linghe and Wang, Yue and others},
  journal={arXiv preprint arXiv:2602.11858},
  year={2026}
}

@article{bai2025qwen3vl,
  title={{Qwen3-VL} Technical Report},
  author={Bai, Shuai and Cai, Yuxuan and Chen, Ruizhe and Chen, Keqin and Chen, Xionghui and Cheng, Zesen and others},
  journal={arXiv preprint arXiv:2511.21631},
  year={2025}
}

@article{zheng2025deepeyes,
  title={Deepeyes: Incentivizing" thinking with images" via reinforcement learning},
  author={Zheng, Ziwei and Yang, Michael and Hong, Jack and Zhao, Chenxiao and Xu, Guohai and Yang, Le and Shen, Chao and Yu, Xing},
  journal={arXiv preprint arXiv:2505.14362},
  year={2025}
}

@article{hong2025deepeyesv2,
  title={Deepeyesv2: Toward agentic multimodal model},
  author={Hong, Jack and Zhao, Chenxiao and Zhu, ChengLin and Lu, Weiheng and Xu, Guohai and Yu, Xing},
  journal={arXiv preprint arXiv:2511.05271},
  year={2025}
}

@article{zhang2025thyme,
  title={Thyme: Think beyond images},
  author={Zhang, Yi-Fan and Lu, Xingyu and Yin, Shukang and Fu, Chaoyou and Chen, Wei and Hu, Xiao and Wen, Bin and Jiang, Kaiyu and Liu, Changyi and Zhang, Tianke and others},
  journal={arXiv preprint arXiv:2508.11630},
  year={2025}
}

@article{chng2025sensenova,
  title={SenseNova-MARS: Empowering Multimodal Agentic Reasoning and Search via Reinforcement Learning},
  author={Chng, Yong Xien and Hu, Tao and Tong, Wenwen and Li, Xueheng and Chen, Jiandong and Yu, Haojia and Lu, Jiefan and Guo, Hewei and Deng, Hanming and Xie, Chengjun and others},
  journal={arXiv preprint arXiv:2512.24330},
  year={2025}
}

@misc{zhang2026instruction,
      title={Instruction Anchor: Dissecting the Mechanistic Dynamics of Modality Arbitration}, 
      author={Yu Zhang and Mufan Xu and Xuefeng Bai and Kehai Chen and Pengfei Zhang and Yang Xiang and Min Zhang},
      year={2026},
      eprint={2602.03677},
      archivePrefix={arXiv},
      primaryClass={cs.CL},
      url={https://arxiv.org/abs/2602.03677}, 
}

@article{zhang2025evaluating,
  title={Evaluating and steering modality preferences in multimodal large language model},
  author={Zhang, Yu and Ma, Jinlong and Hou, Yongshuai and Bai, Xuefeng and Chen, Kehai and Xiang, Yang and Yu, Jun and Zhang, Min},
  journal={arXiv preprint arXiv:2505.20977},
  year={2025}
}

@inproceedings{park2025generalizing,
  title={Generalizing from {SIMPLE} to {HARD} Visual Reasoning: Can We Mitigate Modality Imbalance in {VLM}s?},
  author={Park, Simon and Panigrahi, Abhishek and Cheng, Yun and Yu, Dingli and Goyal, Anirudh and Arora, Sanjeev},
  booktitle={Proceedings of the 42nd International Conference on Machine Learning},
  pages={48192--48244},
  year={2025},
  volume={267},
  series={Proceedings of Machine Learning Research},
  publisher={PMLR},
  url={https://proceedings.mlr.press/v267/park25j.html}
}

@article{hong2025glm,
  title={Glm-4.5 v and glm-4.1 v-thinking: Towards versatile multimodal reasoning with scalable reinforcement learning},
  author={Hong, Wenyi and Yu, Wenmeng and Gu, Xiaotao and Wang, Guo and Gan, Guobing and Tang, Haomiao and Cheng, Jiale and Qi, Ji and Ji, Junhui and Pan, Lihang and others},
  journal={arXiv preprint arXiv:2507.01006},
  year={2025}
}

@misc{qwen3.5,
    title  = {{Qwen3.5}: Towards Native Multimodal Agents},
    author = {{Qwen Team}},
    month  = {February},
    year   = {2026},
    url    = {https://qwen.ai/blog?id=qwen3.5}
}

@inproceedings{wu2024v,
  title={V*: Guided visual search as a core mechanism in multimodal llms},
  author={Wu, Penghao and Xie, Saining},
  booktitle={Proceedings of the IEEE/CVF Conference on Computer Vision and Pattern Recognition},
  pages={13084--13094},
  year={2024}
}

@inproceedings{wang2025divide,
  title={Divide, conquer and combine: A training-free framework for high-resolution image perception in multimodal large language models},
  author={Wang, Wenbin and Ding, Liang and Zeng, Minyan and Zhou, Xiabin and Shen, Li and Luo, Yong and Yu, Wei and Tao, Dacheng},
  booktitle={Proceedings of the AAAI Conference on Artificial Intelligence},
  volume={39},
  pages={7907--7915},
  year={2025}
}

@article{zhang2024mme,
  title={Mme-realworld: Could your multimodal llm challenge high-resolution real-world scenarios that are difficult for humans?},
  author={Zhang, Yi-Fan and Zhang, Huanyu and Tian, Haochen and Fu, Chaoyou and Zhang, Shuangqing and Wu, Junfei and Li, Feng and Wang, Kun and Wen, Qingsong and Zhang, Zhang and others},
  journal={arXiv preprint arXiv:2408.13257},
  year={2024}
}

@article{yu2025minicpm,
  title={Minicpm-v 4.5: Cooking efficient mllms via architecture, data, and training recipe},
  author={Yu, Tianyu and Wang, Zefan and Wang, Chongyi and Huang, Fuwei and Ma, Wenshuo and He, Zhihui and Cai, Tianchi and Chen, Weize and Huang, Yuxiang and Zhao, Yuanqian and others},
  journal={arXiv preprint arXiv:2509.18154},
  year={2025}
}

@misc{coreteam2025mimovltechnicalreport,
      title={MiMo-VL Technical Report},
      author={LLM-Core-Team Xiaomi},
      year={2025},
      eprint={2506.03569},
      archivePrefix={arXiv},
      primaryClass={cs.CL},
      url={https://arxiv.org/abs/2506.03569},
}

@misc{team2026kimi26,
  author        = {{Team, Kimi}},
  title        = {Kimi K2.6: From Code to Creation, From One to Many},
  howpublished = {\url{https://www.kimi.com/ai-models/kimi-k2-6/}},
  year         = {2026}
}

@misc{google2026gemini3,
  author        = {{Google}},
  title        = {Gemini 3.1 Pro},
  howpublished = {\url{https://deepmind.google/models/model-cards/gemini-3-1-pro/}},
  year         = {2026}
}

@misc{google2026gemini35,
  author        = {{Google}},
  title        = {Gemini 3.5 Flash},
  howpublished = {\url{https://deepmind.google/models/model-cards/gemini-3-5-flash/}},
  year         = {2026}
}

@misc{openai2025gpt52,
  author        = {{OpenAI}},
  title        = {Introducing GPT-5.2},
  howpublished = {\url{https://openai.com/index/introducing-gpt-5-2/}},
  year         = {2025}
}

@misc{openai2026gpt54,
  author        = {{OpenAI}},
  title        = {Introducing GPT-5.4},
  howpublished = {\url{https://openai.com/index/introducing-gpt-5-4/}},
  year         = {2026}
}

@article{yuan2026visionopd,
  title={Vision-OPD: Learning to See Fine Details for Multimodal LLMs via On-Policy Self-Distillation},
  author={Yuan, Qianhao and Lou, Jie and Yu, Xing and Lin, Hongyu and Sun, Le and Han, Xianpei and Lu, Yaojie},
  journal={arXiv preprint arXiv:2605.18740},
  year={2026}
}

@article{wang2026seeing,
  title={Seeing Before Reasoning: Decoupling Perception and Reasoning for Shortcut-Resilient Multimodal On-Policy Self-Distillation},
  author={Wang, Sihan and Liu, Xiyao and Liu, Lianqing and Han, Zhi},
  journal={arXiv preprint arXiv:2606.19120},
  year={2026}
}

@inproceedings{bi-etal-2025-llava,
    title = "{LL}a{VA} Steering: Visual Instruction Tuning with 500x Fewer Parameters through Modality Linear Representation-Steering",
    author = "Bi, Jinhe  and
      Wang, Yujun  and
      Chen, Haokun  and
      Xiao, Xun  and
      Hecker, Artur  and
      Tresp, Volker  and
      Ma, Yunpu",
    editor = "Che, Wanxiang  and
      Nabende, Joyce  and
      Shutova, Ekaterina  and
      Pilehvar, Mohammad Taher",
    booktitle = "Proceedings of the 63rd Annual Meeting of the Association for Computational Linguistics (Volume 1: Long Papers)",
    month = jul,
    year = "2025",
    address = "Vienna, Austria",
    publisher = "Association for Computational Linguistics",
    url = "https://aclanthology.org/2025.acl-long.739/",
    pages = "15230--15250",
    ISBN = "979-8-89176-251-0"
}

@article{li2026visualopsd,
  title={Visual-OPSD: Cross-Modal On-Policy Self-Distillation for Efficient Unified Multimodal Reasoning},
  author={Li, Pengyu and Gao, Zhitao and Zhang, Lingling and Huang, Muye and Li, Yuanming and Xu, Fangzhi and Liu, Jun},
  journal={arXiv preprint arXiv:2606.18974},
  year={2026}
}

@article{liu2026visualadvantage,
  title={Visual-Advantage On-Policy Distillation for Vision-Language Models},
  author={Liu, Ruiqi and Lv, Xiaolei and Li, Gengsheng and Zhu, Ximo and Wang, Zhiheng and Zhang, Zhengbo and Chen, Junkai and Li, Zhiheng and Li, Bo and Gao, Jun and Wu, Shu},
  journal={arXiv preprint arXiv:2605.21924},
  year={2026}
}

@article{liang2026visualcontrastive,
  title={Visual Contrastive Self-Distillation},
  author={Liang, Yijun and Tian, Yunjie and Li, Yijiang and Jia, Yuqi and Huang, Furong and Zhou, Tianyi and Fu, Di},
  journal={arXiv preprint arXiv:2607.21556},
  year={2026}
}

@article{Wu2025WhenMI,
  title={When More is Less: Understanding Chain-of-Thought Length in LLMs},
  author={Yuyang Wu and Yifei Wang and Tianqi Du and Stefanie Jegelka and Yisen Wang},
  journal={ArXiv},
  year={2025},
  volume={abs/2502.07266},
  url={https://api.semanticscholar.org/CorpusID:276259519}
}

@misc{
hassid2026dont,
title={Don't Overthink it. Preferring Shorter Thinking Chains for Improved {LLM} Reasoning},
author={Michael Hassid and Gabriel Synnaeve and Yossi Adi and Roy Schwartz},
year={2026},
url={https://openreview.net/forum?id=nhUlA8iMkD}
}

@article{Bi2025PRISMSI,
  title={PRISM: Self-Pruning Intrinsic Selection Method for Training-Free Multimodal Data Selection},
  author={Jinhe Bi and Yifan Wang and Danqi Yan and Xun Xiao and Artur Hecker and Volker Tresp and Yunpu Ma},
  journal={ArXiv},
  year={2025},
  volume={abs/2502.12119},
  url={https://api.semanticscholar.org/CorpusID:276421326}
}

@article{Wang_Bi_Pirk_Ma_2026, title={ASCD: Attention-Steerable Contrastive Decoding for Reducing Hallucination in MLLM}, volume={40}, url={https://ojs.aaai.org/index.php/AAAI/article/view/38000}, DOI={10.1609/aaai.v40i12.38000}, abstractNote={Multimodal large language models (MLLMs) frequently hallucinate by over-committing to spurious visual cues. Prior remedies–Visual and Instruction Contrastive Decoding (VCD, ICD)–mitigate this issue, yet the mechanism remains opaque. We first empirically show that their improvements systematically coincide with redistributions of cross-modal attention. Building on this insight, we propose Attention-Steerable Contrastive Decoding (ASCD), which directly steers the attention scores during decoding. ASCD combines (i) positive steering, which amplifies automatically mined text-centric heads–stable within a model and robust across domains–with (ii) negative steering, which dampens on-the-fly identified critical visual tokens. The method incurs negligible runtime/memory overhead and requires no additional training. Across five MLLM backbones and three decoding schemes, ASCD reduces hallucination on POPE, CHAIR, and MMHal-Bench by up to 38.2% while improving accuracy on standard VQA benchmarks, including MMMU, MM-VET, ScienceQA, TextVQA, and GQA. These results position attention steering as a simple, model-agnostic, and principled route to safer, more faithful multimodal generation.}, number={12}, journal={Proceedings of the AAAI Conference on Artificial Intelligence}, author={Wang, Yujun and , Aniri and Bi, Jinhe and Pirk, Soren and Ma, Yunpu}, year={2026}, month={Mar.}, pages={10306–10314} }

@article{zhang2023spot,
  title={SPOT! Revisiting Video-Language Models for Event Understanding},
  author={Zhang, Gengyuan and Bi, Jinhe and Gu, Jindong and Chen, Yanyu and Tresp, Volker},
  journal={arXiv preprint arXiv:2311.12919},
  year={2023}
}

@misc{peng2025visualinputcompressedvisual,
      title={Can Visual Input Be Compressed? A Visual Token Compression Benchmark for Large Multimodal Models}, 
      author={Tianfan Peng and Yuntao Du and Pengzhou Ji and Shijie Dong and Kailin Jiang and Mingchuan Ma and Yijun Tian and Jinhe Bi and Qian Li and Wei Du and Feng Xiao and Lizhen Cui},
      year={2025},
      eprint={2511.02650},
      archivePrefix={arXiv},
      primaryClass={cs.CV},
      url={https://arxiv.org/abs/2511.02650}, 
}

@misc{jiang2025minedprobingupdatingmultimodal,
      title={MINED: Probing and Updating with Multimodal Time-Sensitive Knowledge for Large Multimodal Models}, 
      author={Kailin Jiang and Ning Jiang and Yuntao Du and Yuchen Ren and Yuchen Li and Yifan Gao and Jinhe Bi and Yunpu Ma and Qingqing Liu and Xianhao Wang and Yifan Jia and Hongbo Jiang and Yaocong Hu and Bin Li and Lei Liu},
      year={2025},
      eprint={2510.19457},
      archivePrefix={arXiv},
      primaryClass={cs.CL},
      url={https://arxiv.org/abs/2510.19457}, 
}

@misc{jiang2025koreenhancingknowledgeinjection,
      title={KORE: Enhancing Knowledge Injection for Large Multimodal Models via Knowledge-Oriented Augmentations and Constraints}, 
      author={Kailin Jiang and Hongbo Jiang and Ning Jiang and Zhi Gao and Jinhe Bi and Yuchen Ren and Bin Li and Yuntao Du and Lei Liu and Qing Li},
      year={2025},
      eprint={2510.19316},
      archivePrefix={arXiv},
      primaryClass={cs.CL},
      url={https://arxiv.org/abs/2510.19316}, 
}

@inproceedings{
bi2026echorl,
title={Echo{RL}: Reinforcement Learning via Rollout Echoing},
author = {Jinhe Bi and Aniri and Minglai Yang and Xingcheng Zhou and Wenke Huang and Sikuan Yan and Yujun Wang and Zixuan Cao and Michael Färber and Xun Xiao and Volker Tresp and Yunpu Ma}
,
booktitle={Forty-third International Conference on Machine Learning},
year={2026},
url={https://openreview.net/forum?id=A6az59SGtF}
}

@article{Bi2025CoTKineticsAT,
  title={CoT-Kinetics: A Theoretical Modeling Assessing LRM Reasoning Process},
  author={Jinhe Bi and Danqi Yan and Yifan Wang and Wenke Huang and Haokun Chen and Guancheng Wan and Mang Ye and Xun Xiao and Hinrich Schuetze and Volker Tresp and Yunpu Ma},
  journal={ArXiv},
  year={2025},
  volume={abs/2505.13408},
  url={https://api.semanticscholar.org/CorpusID:278769227}
}

@misc{huang2025loongsynthesizelongchainofthoughts,
      title={Loong: Synthesize Long Chain-of-Thoughts at Scale through Verifiers}, 
      author={Xingyue Huang and Rishabh and Gregor Franke and Ziyi Yang and Jiamu Bai and Weijie Bai and Jinhe Bi and Zifeng Ding and Yiqun Duan and Chengyu Fan and Wendong Fan and Xin Gao and Ruohao Guo and Yuan He and Zhuangzhuang He and Xianglong Hu and Neil Johnson and Bowen Li and Fangru Lin and Siyu Lin and Tong Liu and Yunpu Ma and Hao Shen and Hao Sun and Beibei Wang and Fangyijie Wang and Hao Wang and Haoran Wang and Yang Wang and Yifeng Wang and Zhaowei Wang and Ziyang Wang and Yifan Wu and Zikai Xiao and Chengxing Xie and Fan Yang and Junxiao Yang and Qianshuo Ye and Ziyu Ye and Guangtao Zeng and Yuwen Ebony Zhang and Zeyu Zhang and Zihao Zhu and Bernard Ghanem and Philip Torr and Guohao Li},
      year={2025},
      eprint={2509.03059},
      archivePrefix={arXiv},
      primaryClass={cs.LG},
      url={https://arxiv.org/abs/2509.03059}, 
}

@inproceedings{ma-etal-2026-self,
    title = "Self-Evolving Multi-Agent Systems via Textual Backpropagation",
    author = "Ma, Xiaowen  and
      Ma, Yunpu  and
      Lin, Chenyang  and
      Yan, Sikuan  and
      Bi, Jinhe  and
      Cao, Zixuan  and
      Tian, Yijun  and
      Tresp, Volker  and
      Schuetze, Hinrich",
    editor = "Liakata, Maria  and
      Moreira, Viviane P.  and
      Zhang, Jiajun  and
      Jurgens, David",
    booktitle = "Findings of the {A}ssociation for {C}omputational {L}inguistics: {ACL} 2026",
    month = jul,
    year = "2026",
    address = "San Diego, California, United States",
    publisher = "Association for Computational Linguistics",
    url = "https://aclanthology.org/2026.findings-acl.483/",
    doi = "10.18653/v1/2026.findings-acl.483",
    pages = "9918--9951",
    ISBN = "979-8-89176-395-1"
}

@misc{zhao2026nl2codestructuredsurveymultimodal,
      title={Beyond NL2Code: A Structured Survey of Multimodal Code Intelligence}, 
      author={Xuanle Zhao and Qiushi Sun and Jingyu Xiao and Xuexin Liu and Haoyue Yang and Qiaosheng Chen and Xianzhen Luo and Jing Huang and Yufeng Zhong and Lei Chen and Shuai Fu and Zhenlin Wei and Jinhe Bi and Lei Jiang and Haibo Qiu and Siqi Yang and Peng Shi and Jian Hu and Zhixiong Zeng},
      year={2026},
      eprint={2606.15932},
      archivePrefix={arXiv},
      primaryClass={cs.CL},
      url={https://arxiv.org/abs/2606.15932}, 
}

@misc{yu2026dopddualonpolicydistillation,
      title={DOPD: Dual On-policy Distillation}, 
      author={Xinlei Yu and Gen Li and Qingyi Si and Guibin Zhang and Yuqi Xu and Congcong Wang and Shuai Dong and Kaiwen Tuo and Xiangyu Zeng and Kaituo Feng and Qunzhong Wang and Yang Shi and Xiaobin Hu and Xiangyu Yue and Jiaqi Wang and Shuicheng Yan},
      year={2026},
      eprint={2606.30626},
      archivePrefix={arXiv},
      primaryClass={cs.AI},
      url={https://arxiv.org/abs/2606.30626}, 
}

@misc{bi2026reflectrllearninggoldennegative,
      title={ReflectRL: Learning from Golden Negative Trajectories via Reflective-to-Direct Reasoning}, 
      author={Jinhe Bi and Chennan Zhou and Zengjie Jin and Aniri and Shuo Lu and Wenke Huang and Hu Cao and Xun Xiao and Zhihong Zhu and Volker Tresp and Fei Shen and Yunpu Ma and Tat-Seng Chua},
      year={2026},
      eprint={2608.03972},
      archivePrefix={arXiv},
      primaryClass={cs.AI},
      url={https://arxiv.org/abs/2608.03972}, 
}

\beginappendix

\section{Related Work}

\subsection{From SFT to On-Policy Self-Distillation}

Supervised fine-tuning (SFT) trains a model on reference responses with teacher-forced prefixes. The objective is simple and stable, yet the supervision remains tied to fixed reference trajectories rather than the states produced by the current student. Knowledge distillation changes the target from a single reference token to a teacher distribution \cite{hinton2015distilling,kim2016sequence,Bi2025CoTKineticsAT,bi2026echorl}. On-Policy Distillation (OPD) brings this distributional supervision to student-generated trajectories: the student first samples its own response, and the teacher evaluates the prefixes actually visited by the student \cite{agarwal2024policy,lu2025onpolicydistillation}. Compared with SFT, OPD exposes model errors on the model's own rollouts and provides token-level feedback at those states.

On-Policy Self-Distillation (OPSD) keeps the on-policy structure while replacing the external teacher with a detached copy of the same model \cite{zhao2026self}. At each token position, the teacher scores the same student-generated prefix, and the distillation loss updates the student distribution at that visited state. This makes the teacher condition central: the trajectory determines where supervision is applied, while the condition determines what signal is distilled. In MLLMs, the condition includes the image input because each prediction depends on the Original Image, textual query, and autoregressive prefix. Recent visual OPD and OPSD methods use transformed visual inputs or visual reasoning signals to make teacher supervision more informative \cite{yuan2026visionopd,li2026visualopsd,liu2026visualadvantage,liang2026visualcontrastive,wang2026seeing}. OPD-V adds a Modality Balance criterion before applying self-distillation, favoring teacher support that reflects stronger use of visual evidence rather than confidence driven mainly by textual context.

\FloatBarrier
\begin{algorithm}[h]
\caption{OPD-V Training}
\label{alg:opdv}
\begin{algorithmic}[1]
\REQUIRE Training set \(\mathcal{D}\), student model \(p_\theta\), rollout count \(n\), EMA update rate \(\tau\)
\ENSURE Updated student parameters \(\theta\)
\STATE Initialize teacher parameters \(\bar{\theta}\leftarrow\theta\)
\FOR{each training step}
    \STATE Sample a mini-batch \(\mathcal{B}\subset\mathcal{D}\)
    \STATE Initialize mini-batch loss \(\mathcal{L}_{\mathcal{B}}\leftarrow 0\)
    \FOR{each \((I,x)\in\mathcal{B}\)}
        \STATE Obtain \(I^{\mathrm{zoom}}\) from the training data
        \STATE \(I^{\mathrm{mask}}\leftarrow\operatorname{Mask}(I^{\mathrm{zoom}})\)
        \FOR{\(j=1,\ldots,n\)}
            \STATE Sample rollout \(y=(y_1,\ldots,y_T)\sim p_\theta(\cdot\mid I,x)\)
            \STATE Initialize rollout loss \(\mathcal{L}_{y}\leftarrow 0\)
            \FOR{each \(t\in\mathcal{T}_y\)}
                \STATE Set \(r_t\leftarrow 1\) for generated response tokens
                \STATE Compute \(p_{\theta,t}\), \(q_t^+\), and \(q_t^-\)
                \STATE \(\delta_t^{\mathrm{MB}}\leftarrow\log q_t^+(y_t)-\log q_t^-(y_t)\)
                \IF{\(\delta_t^{\mathrm{MB}}>0\) and \(r_t=1\)}
                    \STATE \(D_t\leftarrow D_{\mathrm{JS}}(q_t^+,p_{\theta,t})\)
                    \STATE \(\mathcal{L}_{y}\leftarrow\mathcal{L}_{y}+\delta_t^{\mathrm{MB}}D_t\)
                \ENDIF
            \ENDFOR
            \STATE \(\mathcal{L}_{\mathcal{B}}\leftarrow\mathcal{L}_{\mathcal{B}}+\mathcal{L}_{y}/\sum_{t\in\mathcal{T}_y}r_t\)
        \ENDFOR
    \ENDFOR
    \STATE Update \(\theta\) by minimizing \(\mathcal{L}_{\mathcal{B}}/(n|\mathcal{B}|)\)
    \STATE Update teacher parameters \(\bar{\theta}\leftarrow(1-\tau)\bar{\theta}+\tau\theta\)
\ENDFOR
\end{algorithmic}
\end{algorithm}

\subsection{Modality Imbalance in MLLMs}

Modality Imbalance denotes the tendency of an MLLM to depend disproportionately on textual context relative to visual input during response generation. Prior studies show that textual information can dominate multimodal prediction even when the query requires visual evidence \cite{bi-etal-2025-llava,park2025generalizing,zhang2025evaluating,zhang2026instruction,Bi2025PRISMSI,peng2025visualinputcompressedvisual,Wang_Bi_Pirk_Ma_2026,jiang2025minedprobingupdatingmultimodal,jiang2025koreenhancingknowledgeinjection}. This creates a training-time ambiguity for OPSD. A confident teacher distribution may reflect visual evidence, language priors, or the textual query, so the relevant visual content can remain underused even when token-level supervision appears strong.

Once OPSD uses teacher distributions as token-level supervision, this ambiguity affects token selection. Selecting tokens by teacher confidence alone can preserve supervision driven mainly by textual context, consistent with the shortcut supervision observed in multimodal OPSD \cite{wang2026seeing}. OPD-V makes Modality Balance part of the self-distillation criterion, so token selection depends on whether teacher support reflects stronger use of visual evidence rather than confidence alone.

\begin{figure*}[t]
\centering
\includegraphics[width=\textwidth]{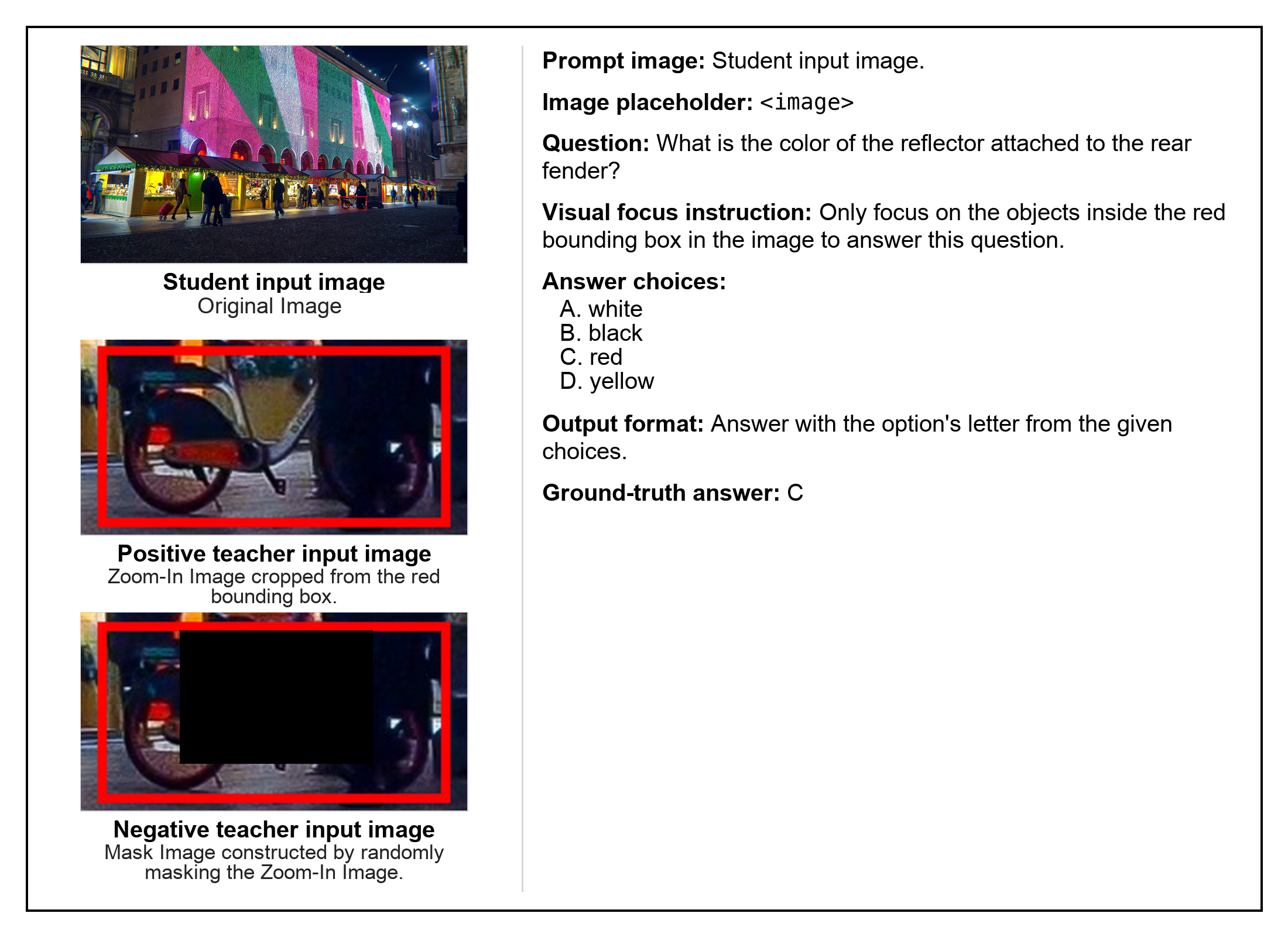}
\caption{Training example for a rear-fender reflector question.}
\label{fig:appendix_prompt_examples}
\end{figure*}

\begin{table*}[t]
\centering
\small
\setlength{\tabcolsep}{4pt}
\begin{tabular}{@{}lcp{0.60\textwidth}@{}}
\toprule
\textbf{Benchmark} & \textbf{Samples} & \textbf{Scoring Rule} \\
\midrule
V* Bench & 191 & Rule-based answer matching is applied first; unresolved responses are compared with the verified answer by a Yes/No judge. \\
ZoomBench & 845 & Generic answer matching handles normalized responses first; unresolved cases use the verified-answer Yes/No judge. \\
HR-Bench 4K & 800 & Multiple-choice matching checks direct answer matches and option-letter matches before the Yes/No judge handles unresolved responses. \\
HR-Bench 8K & 800 & Multiple-choice matching follows the HR-Bench 4K rule. \\
MME-RealWorld EN & 23609 & Multiple-choice matching and the Yes/No judge are applied to the English subset. \\
MME-RealWorld CN & 5917 & The Chinese subset uses the MME-RealWorld multiple-choice scoring rule. \\
\bottomrule
\end{tabular}
\caption{Evaluation protocol for benchmarks.}
\label{tab:appendix_eval_protocol}
\end{table*}

\section{OPD-V Algorithm}

In algorithm~\ref{alg:opdv} for OPD-V: each prompt first yields \(n\) on-policy rollouts, and teacher scoring is then aligned to the same visited prefixes. At token position \(t\), \(p_{\theta,t}\), \(q_t^+\), and \(q_t^-\) are calculated under the shared prefix. The Positive Teacher with the Zoom-In Image provides the Jensen--Shannon distillation target, while the Negative Teacher with the Mask Image contributes the paired comparison for the Modality-Balance Trust Region. Only generated response tokens with \(r_t=1\) enter the loss, and the detached teacher is updated by EMA after the student update.

\subsection{Loss and Top-\(K\) Distillation Details}

Token selection uses the Modality-Balance Logits Margin \(\delta_t^{\mathrm{MB}}\) on the student-generated token \(y_t\). A positive Logits Margin indicates that the same token receives more support from the Positive Teacher than from the Negative Teacher, and these positions form the Modality-Balance Trust Region. Within this region, \(\delta_t^{\mathrm{MB}}\) scales the selected on-policy tokens during self-distillation. The distillation target remains distributional: \(D_t\) compares the Positive Teacher with the Zoom-In Image and student distributions, rather than only the probabilities assigned to \(y_t\).

Full-vocabulary logit distillation is memory intensive for MLLMs, so the implementation first takes the student top-\(K\) token set at each response position:
\begin{equation}
    \mathcal{K}_t=\operatorname{TopK}\!\left(p_{\theta,t},K\right),
    \qquad K=100 .
\end{equation}
The Positive Teacher with the Zoom-In Image is evaluated on the same set \(\mathcal{K}_t\). For each distribution \(s\in\{p_{\theta,t},q_t^+\}\), the top-\(K\) probabilities are retained and one tail bucket stores the remaining vocabulary mass:
\begin{equation}
    s_t^{\mathrm{tail}}
    =
    1-\sum_{v\in\mathcal{K}_t}s(v).
\end{equation}
The \((K+1)\)-dimensional form preserves the probability assigned to tokens outside \(\mathcal{K}_t\) without materializing the full vocabulary. With \(q_t^+\) and \(p_{\theta,t}\) represented by their tail-adjusted top-\(K\) distributions, per-token distillation term expands as
\begin{equation}
\begin{aligned}
D_t
&=
D_{\mathrm{JS}}\!\left(q_t^+,p_{\theta,t}\right)
\\
&=
\frac{1}{2}D_{\mathrm{KL}}\!\left(q_t^+\,\|\,m_t\right)
+
\frac{1}{2}D_{\mathrm{KL}}\!\left(p_{\theta,t}\,\|\,m_t\right),
\\
m_t
&=
\frac{1}{2}\left(q_t^+ + p_{\theta,t}\right).
\end{aligned}
\label{eq:appendix_topk_jsd}
\end{equation}
The Negative Teacher with the Mask Image enters only through \(\delta_t^{\mathrm{MB}}\), while the Positive Teacher with the Zoom-In Image remains the distillation target.

Loss reduction follows the valid-token count rather than the sum of margin weights. The normalization stays tied to generated response tokens, while \(\delta_t^{\mathrm{MB}}\) controls the strength of the selected distillation signal.

\subsection{Image Operations}

OPD-V changes the teacher image input while keeping the textual query, on-policy prefix, and teacher parameters fixed. The Zoom-In Image and Mask Image are derived from the same Original Image, which includes the target-region cue. The Positive Teacher with the Zoom-In Image receives \(I^{\mathrm{zoom}}\). In the experiments, \(I^{\mathrm{zoom}}\) is taken from the evidence-centered crop supplied with the Vision-OPD training data \cite{yuan2026visionopd}. The crop is defined by the bounding box around the task-relevant region and is magnified before image processing, increasing the prominence of the visual evidence required by the query.

The Negative Teacher with the Mask Image receives \(I^{\mathrm{mask}}\), computed as \(I^{\mathrm{mask}}=\operatorname{Mask}(I^{\mathrm{zoom}})\). The Mask Image is constructed from the evidence-centered cropped Original Image \(I^{\mathrm{zoom}}\) by replacing a randomly selected rectangular region with black pixels before image processing, forming a visually weakened condition while preserving most image context. The Positive Teacher with the Zoom-In Image and the Negative Teacher with the Mask Image share the same textual query, on-policy prefix, and EMA parameters, so their distributional difference reflects the image operation applied to the teacher input.

\begin{table*}[t]
\centering
\small
\setlength{\tabcolsep}{3pt}
\begin{tabular}{@{}p{0.19\textwidth}cp{0.24\textwidth}p{0.43\textwidth}@{}}
\toprule
\textbf{Backbone} & \textbf{Params} & \textbf{Model Identifier} & \textbf{VLM and Perception Stack} \\
\midrule
Qwen3.5-4B & 4B & Qwen/Qwen3.5-4B & Qwen3.5 multimodal backbone with early vision--text fusion and a 3:1 hybrid language backbone that interleaves Gated DeltaNet linear-attention layers with full-attention layers. \\
Qwen3.5-9B & 9B & Qwen/Qwen3.5-9B & Qwen3.5 multimodal backbone with the same vision--text fusion design as Qwen3.5-4B, scaled to the 9B parameter setting. \\
Qwen3-VL-4B-Instruct & 4B & Qwen/Qwen3-VL-4B-Instruct & Qwen3-VL vision-language stack with Interleaved-MRoPE and DeepStack, which injects multi-level ViT features into the language backbone. \\
Qwen3-VL-8B-Instruct & 8B & Qwen/Qwen3-VL-8B-Instruct & Qwen3-VL vision-language stack with the same Interleaved-MRoPE and DeepStack design as Qwen3-VL-4B-Instruct, scaled to the 8B parameter setting. \\
\bottomrule
\end{tabular}
\caption{Backbone configurations used in the OPD-V experiments.}
\label{tab:appendix_model_config}
\end{table*}

\begin{table*}[t]
\centering
\small
\setlength{\tabcolsep}{4pt}
\begin{tabular}{@{}lll@{}}
\toprule
\textbf{Setting} & \textbf{Value} & \textbf{Role} \\
\midrule
GPU allocation & 1 node, 4 H200 GPUs & Training worker allocation \\
Training batch size & 48 & Mini-batch size per training step \\
Actor update mini-batch size & 48 & Mini-batch size for actor update \\
Rollouts per prompt & 8 & Number of on-policy responses sampled per prompt \\
Maximum prompt length & 8192 tokens & Prompt truncation limit \\
Maximum response length & 1024 tokens & On-policy generation limit \\
Training epochs & 1 & Number of passes over the training set \\
Policy loss mode & OPD-V & Self-distillation objective used for actor update \\
Distillation objective & Jensen--Shannon divergence & Token-level distribution matching \\
Top-\(K\) size & 100 & Candidate set for distillation distributions \\
Teacher parameter update & EMA of student weights & Detached teacher update rule \\
Teacher EMA update rate & 0.05 & Student-weight coefficient in the EMA update \\
Teacher construction & Always on & Teacher inputs are built at every training step \\
Positive teacher image field & bbox\_images & Source of the Zoom-In Image \\
Negative teacher image operation & Mask Image & Random rectangular masking of the Zoom-In Image \\
Rollout correction mode & Token & Token-level rollout correction \\
Rollout correction threshold & 2.0 & Clipping threshold for rollout correction \\
Checkpoint frequency & \(-1\) & No default checkpoint saving during training \\
Validation frequency & \(-1\) & No default validation during training \\
Logging mode & Offline W\&B & Logs are synced after training \\
\bottomrule
\end{tabular}
\caption{Training hyperparameters for OPD-V.}
\label{tab:appendix_hyperparameters}
\end{table*}

\section{Training Prompt Example}

Figure~\ref{fig:appendix_prompt_examples} presents one training sample in the same field structure used by the prompt: prompt image, image placeholder, question, visual focus instruction, answer choices, output format, and ground-truth answer. The left column marks which image condition is paired with the shared textual prompt in the student and teacher passes.

\section{Model Configuration Details}

Table~\ref{tab:appendix_model_config} separates the four MLLM backbones by model identifier, parameter scale, and VLM/perception stack. This makes the comparison across Qwen3.5 and Qwen3-VL explicit, including the early vision--text fusion design in Qwen3.5 and the Interleaved-MRoPE with DeepStack design in Qwen3-VL.

The training split contains \(6241\) visual reasoning samples from the dataset \cite{yuan2026visionopd}. Each sample provides the Original Image, textual query, verified target, and the dataset field used as \(I^{\mathrm{zoom}}\). The Mask Image is generated from the evidence-centered cropped Original Image \(I^{\mathrm{zoom}}\) during training, and evaluation uses only the Original Image and textual query.

\section{Evaluation Protocol Details}

Evaluation is run in non-thinking mode. Each response is normalized by deterministic matching before unresolved cases are passed to a Yes/No judge. For multiple-choice tasks, direct option matching and first-letter option matching are checked first. Table~\ref{tab:appendix_eval_protocol} pairs each benchmark with its sample count and scoring rule.

The reported Average is the unweighted mean over the six benchmark accuracies:
\begin{equation}
    \mathrm{Average}
    =
    \frac{1}{6}
    \sum_{b\in\mathcal{B}_{\mathrm{eval}}}
    \mathrm{Acc}_{b}.
\end{equation}
This benchmark-level mean differs from sample-weighted overall accuracy, which pools correct predictions over all \(32162\) examples before computing a single percentage.

\section{Hyperparameters and Compute}

Table~\ref{tab:appendix_hyperparameters} records the parameters that control training scale and the teacher--student update schedule. The training batch size and actor update mini-batch size are both 48, so each step uses the sampled mini-batch directly for the actor update. With \(n=8\), every prompt contributes eight on-policy responses before the mini-batch loss is reduced by \(n|\mathcal{B}|\) as in Algorithm~\ref{alg:opdv}.

Sequence limits separate input context from generated trajectories. The 8192-token prompt limit bounds the Original Image, textual query, and prompt formatting, while the 1024-token response limit caps rollout length and the number of token positions scored by the student and teachers. The top-\(K\) parameter is set to \(K=100\), which fixes the candidate set size used by the tail-adjusted distillation distributions described in Section~B.

The EMA update rate \(\tau=0.05\) is the coefficient on the current student weights in the detached teacher update; equivalently, each update retains 0.95 of the previous teacher parameters. Token-level rollout correction uses threshold 2.0 before the actor update. Periodic checkpointing and validation are disabled by the \(-1\) settings, and offline W\&B logging stores run statistics without adding online synchronization during training.

\end{document}